\documentclass[10pt,twocolumn,letterpaper]{article}

\usepackage{wacv}
\usepackage{times}
\usepackage{epsfig}
\usepackage{graphicx}
\usepackage{amsmath}
\usepackage{amssymb}
\usepackage{booktabs}
\usepackage{lipsum}
\usepackage{tikz}
\usetikzlibrary{shapes,arrows}
\usepackage{comment}
\usepackage{caption}
\usepackage{subcaption}

\newcommand*\repeatthanks[1][\value{footnote}]{\footnotemark[#1]}

\usepackage[nolist,nohyperlinks]{acronym}
\acrodef{LDM}[LDM]{Latent Diffusion Model}
\acrodef{VAE}[VAE]{Variational Auto Encoder}
\acrodef{GAN}[GAN]{Generative Adversarial Network}
\acrodef{cGAN}[cGAN]{Conditional Generative Adversarial Network}
\acrodef{DAA}[DAA]{Detect and Avoid}

\def\wacvPaperID{1720} % Enter the WACV Paper ID here

\wacvalgorithmstrack   % Uncomment this line if you are submitting to the Algorithms Track.
\wacvfinalcopy % *** Uncomment this line for the final submission

\ifwacvfinal
\usepackage[breaklinks=true,bookmarks=false]{hyperref}
\else
\usepackage[pagebackref=true,breaklinks=true,colorlinks,bookmarks=false]{hyperref}
\fi

\newcommand\blfootnote[1]{%
  \begingroup
  \renewcommand\thefootnote{}\footnote{#1}%
  \addtocounter{footnote}{-1}%
  \endgroup
}

\begin{document}

\title{Bootstrapping Corner Cases: High-Resolution Inpainting for Safety Critical Detect and Avoid for Automated Flying}

\author{Jonathan Lyhs \thanks{The first two authors contributed equally to this work.}\\
Spleenlab GmbH\\
{\tt\small jonathan.lyhs@spleenlab.ai}
\and
Lars Hinneburg \repeatthanks\\
Spleenlab GmbH\\
{\tt\small lars.hinneburg@spleenlab.ai}
\and
Michael Fischer\\
Spleenlab GmbH\\
{\tt\small michael.fischer@spleenlab.ai}
\and
Florian Ölsner\\
Spleenlab GmbH\\
{\tt\small florian.oelsner@spleenlab.ai}
\and
Stefan Milz\\
Spleenlab GmbH\\
{\tt\small stefan.milz@spleenlab.ai}
\and
Jeremy Tschirner\\
Spleenlab GmbH\\
{\tt\small jeremy.tschirner@spleenlab.ai}
\and
Patrick Mäder\\
Ilmenau University of Technology\\
{\tt\small patrick.maeder@tu-ilmenau.de}
}

\maketitle
\blfootnote{This work was supported by the funded project KI4Flight.}

\thispagestyle{empty}

\begin{abstract}
   Modern machine learning techniques have shown tremendous potential, especially for object detection on camera images. For this reason, they are also used to enable safety-critical automated processes such as autonomous drone flights. We present a study on object detection for Detect and Avoid, a safety critical function for drones that detects air traffic during automated flights for safety reasons. An ill-posed problem is the generation of good and especially large data sets, since detection itself is the corner case. Most models suffer from limited ground truth in raw data, \eg recorded air traffic or frontal flight with a small aircraft. It often leads to poor and critical detection rates. We overcome this problem by using inpainting methods to bootstrap the dataset such that it explicitly contains the corner cases of the raw data. We provide an overview of inpainting methods and generative models and present an example pipeline given a small annotated dataset. We validate our method by generating a high-resolution dataset, which we make publicly available and present it to an independent object detector that was fully trained on real data.
\end{abstract}

\section{Introduction}
A densely collected dataset is essential for the use of machine learning models in a safety-critical domain, such as airborne object detection. Besides the safety aspect, the importance of a sufficient large dataset is a well-studied problem in the field of deep learning. Studies have shown a logarithmic performance increase for image classification based on the volume of the training data \cite{sun2017revisiting}. This stands in contrast to the fact that it is often difficult to obtain data in safety-critical domains.
To mitigate this problem, techniques such as pre-training with synthetic data are used. These approaches are usually based on photo-realistic rendering of multiple scenarios. This brings with it a wealth of automatically generated metadata such as object boxes and segmentation masks. Disadvantages are the previously created effort for scenario modeling and the need for powerful rendering hardware \cite{peng2015learning}.

So far, \acp{GAN} \cite{GAN} have been used for fast and good quality image generation on consumer hardware. Recent work in the field of generative models has reached the state-of-the-art in synthetic image generation. \acp{LDM} \cite{rombach2022high} and discrete \acp{VAE} \cite{kingma2013auto} are used for generating photo-realistic synthetic images. In addition, \acp{LDM} can be run on consumer hardware, making this kind of model available for a wide range of use cases for the first time.

% TODO: better translation here
Therefore we propose a method for generating a synthetic dataset for object detection, based on generative models. We focus on the safety-critical domain of airborne objects. The \ac{DAA} use case is particularly challenging as the systems needs to detect objects at a safe distance. These far away objects will appear extremely small in the images and high-resolution cameras are required to capture them at all.
For efficiency, an inpainting approach is proposed over synthetic generation of the entire scenery.
In particular, we study two different models for this purpose: Pix2Pix \cite{pix2pix2017} a widely used \ac{GAN} and Stable Diffusion\footnote{GitHub: https://github.com/CompVis/stable-diffusion}, which is the reference implementation of \acp{LDM} \cite{rombach2022high} and thus the state-of-the-art in synthetic image generation.
We then propose the pipeline for dataset generation in Section \ref{sec:method}.
The object synthesis processes using Pix2Pix and Stable Diffusion are briefly described in the Sections \ref{sec:synthesis-pix2pix} and \ref{sec:synthesis-sd}.
We substantiate our statements in a series of experiments in Section \ref{sec:experiments}.
The dataset generated in the process is made available for public download\footnote{Generated Dataset:  https://doi.org/10.5281/zenodo.8297255}.

\begin{comment}
Our work can be summarized in the following key contributions:
\begin{enumerate}
    \item We show the application-specific limitations of Stable Diffusion without fine-tuning. 
    \item We propose a method for training Pix2Pix specialized on object inpainting.
    \item We describe a pipeline for generating a high-resolution dataset for object detection.
\end{enumerate}
\end{comment}

\begin{figure*}
  \centering
   \includegraphics[trim=300px 800px 0px 0px,clip,width=\linewidth]{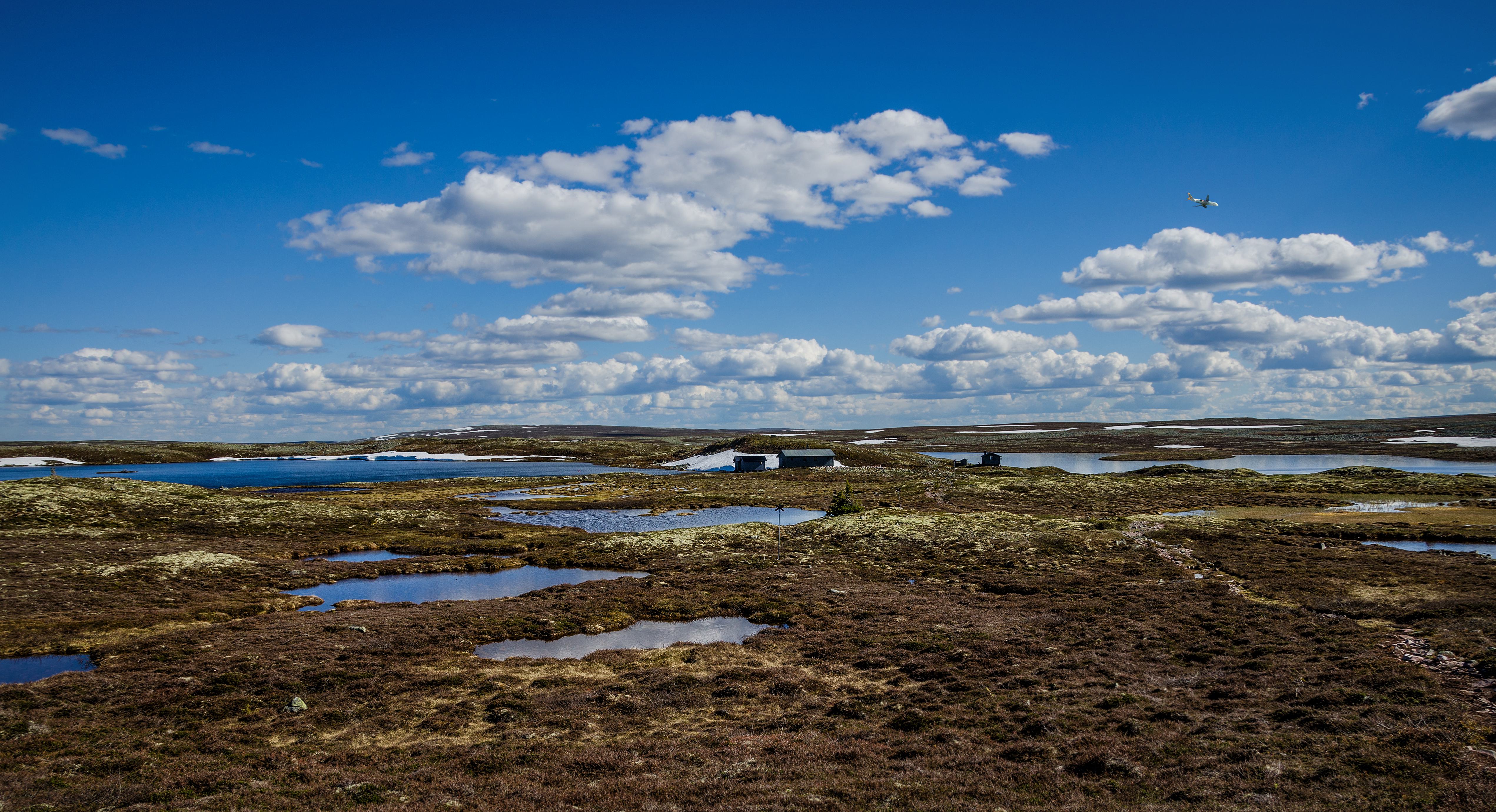}
   \caption{Airborne Object Detection. The high-resolution image shows a small airplane. This is a common situation in Airborne Object Detection for Sense and Avoid functionality. The airplane was generated using our proposed data generation pipeline including ground truth bounding box.}
   \label{fig:aerial-od}
\end{figure*}
    
\section{Related Work}
    
    \textbf{Generative Adversarial Networks}
        \acp{GAN} were introduced in 2014 by Goodfellow \etal \cite{GAN}. They consist of at least two neural networks, a generator and a discriminator. Both networks are trained simultaneously and are configured to work against each other.
        This approach provides efficient sampling of high-resolution images with good perceptual quality. However, the optimization of such models is problematic \cite{rombach2022high}.
        Furthermore with the original \acp{GAN} it is not possible to systematically control the content of the output-image post-training.
        In order to achieve this \acp{cGAN} \cite{cGAN} were introduced. \acp{cGAN} use additional conditions to control the output-image. In an extensive study, Isola \etal \cite{pix2pix2017} show that different types of images, such as photos, segmentation maps, sketches, and more, can be used to condition the model. They also show how the model is able to perform photo inpainting on ground-based urban landscapes, but without the goal of generating specific objects and labeling data. The idea was adapted in a later work \cite{PC2Pix} on an automotive dataset. There the inpainting region is sparsely filled with projected pointcloud data of a car. This additionally conditions shape, position and orientation of the desired object. In contrast, our method specifically targets the airborne use case with small object appearance in high-resolution images. Instead of projected point cloud data, we use segmentation masks as an additional condition for the generator.

    \textbf{VAEs and Flow-Based Models}
        \acp{VAE} \cite{kingma2013auto} and flow-based models \cite{dinh2014nice}\cite{papamakarios2017masked} are two other types of generative models. Both are as efficient as \acp{GAN} in terms of computational resources, but the quality of the images produced has been lower.
        That said, OpenAI's current work, known as DALL-E \cite{ramesh2021zero}, uses a discrete \ac{VAE} to compress images into a computationally more efficient latent space. It also achieves state-of-the-art image synthesis at high-resolution. This work seems to overcome some of the drawbacks that \ac{VAE}-based image generators used to have. It is neither open source nor open science, which limits accessibility for work based on it.
 
    \textbf{Diffusion Probabilistic Models}
        Starting from 2015 different works build up on the concept of diffusion probabilistic models \cite{sohl2015deep}\cite{song2019generative}\cite{ho2020denoising} which are based on a reversible Markov chain of diffusion steps.
        Recent work in this area proposes \acp{LDM} \cite{rombach2022high} which significantly improve accessibility for researchers and end users due to higher inference rates and lower training costs.
        In contrast to DALL-E, the reference implementation of \acp{LDM} is open source and has reached similar performance for high-resolution image synthesis.
        
\section{Method}\label{sec:method}

    \begin{figure*}
      \centering
      %\fbox{\rule{0pt}{2in} \rule{.9\linewidth}{0pt}}
      %\fbox{ \input{images/tikz/data_pipeline.tex}}
      \input{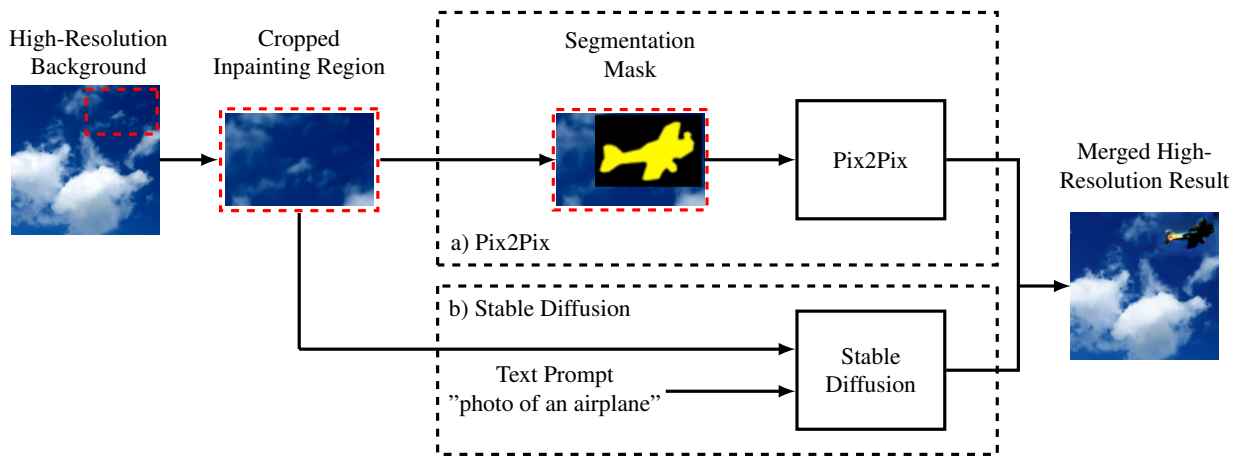}
       \caption{Data Generation Pipeline. Cropping the masked inpainting area from the background is the first step of our proposed pipeline. The cropped image is then passed into the model specific conditioning of the following image synthesis process. The last step is merging the generated image back into the background.}
       \label{fig:data-pipeline}
    \end{figure*}
    
We propose to use generative models to synthetically create images with annotations for the training of object detectors for aircraft in high-resolution imagery. Collection of high quantities of background data is comparably easy. For example a drone equipped with multiple cameras can collect a couple of thousand images for a certain operational domain in a single flight while covering a large area. However, actual encounters with other (especially manned) aircraft like small airplanes, helicopters or hot air balloons are very rare in the wild, expensive to arrange and potentially dangerous. A \ac{DAA} system needs to detect the intruding vehicles at a safe distance such that an avoidance maneuver can be initiated in time. Even large airplanes will appear very small in the high-resolution images when they are far away.
Therefore we propose an inpainting approach that uses existing image data and inserts airborne objects at the desired positions. This is much more efficient at training and inference time and allows the generator to focus on the synthesis of the objects of interest, instead of inventing whole realistically looking landscapes. 

% While inpainting with generative models can create a variety of data the size of the resulting images is still limited. Generating larger images requires more computation-resources and will result in longer inference times. We propose a pipeline to mitigate this problem. By working only on a small part of the image we can keep the inference time and the required memory to a minimum.
% However this also means the size of the inpainted object is more limited.

An overview of our approach is shown in Figure \ref{fig:data-pipeline}. We start with an arbitrary background image that does not contain any objects of interest yet.
We then randomly select a target position at which the object shall be generated. A rectangular region around that point is cropped out of the original image for inpainting. Only that cutout is used during the actual object synthesis stage. After the desired object was drawn into the patch it is fused back into the original high-resolution image. In this work we propose two alternative object synthesis stages, one based on Pix2Pix and the other based on Stable Diffusion which correspond to path a) and b) in Figure \ref{fig:data-pipeline} respectively.

% an inpainting mask to this image. The next step is to crop the masked region from the high-resolution image. Only the cropped image patch is used during the following image synthesis process.
% The subsequent steps depend on the used generative model. For our trained Pix2Pix model a segmentation mask has to be added into the masked region. For inpainting with Stable Diffusion a prompt must be specified for conditioning the synthesis process.
% After the inference step, the result is merged back into the background image. The final image is stored along with the label data. The inpainting mask is used as reference for estimating the bounding box. The object class follows from the conditioning of the generation process.

       % \item start with big image
       % \item crop an area usable by Pix2Pix (256x256)
       % \item add segmentation mask to crop
       % \item inpainting
       % \item calculate the bbox based on the coordinates of the crop and segmentation mask

        %\begin{itemize}
         %   \item creating large images with GANs is expensive 
          %  \item inpainting helps but but is also limited (max Size, Size can't vary)
          %  \item Solution: take a crop from a big image and use in-paint function on the crop only
          %  \item Pro: allows any Image Size, less expensive
          %  \item Con: Object Size might be limited
        %\end{itemize}

\subsection{Object Synthesis using Pix2Pix} \label{sec:synthesis-pix2pix}
    Pix2Pix is a \ac{cGAN} that we condition on the cropped inpainting region. We follow the approach by Milz \etal \cite{PC2Pix} and overlay the inpainting mask in form of a segmentation map directly on the image data. The location of that map is arbitrary in the crop and the size is chosen to be smaller than the full sized image region such that sufficient background context is available to the model. The resulting patch is then passed to the generator network. Ground truth labels can be directly derived from the segmentation map that was used. The minimal enclosing rectangle around the object defines the bounding box and the object class is color coded.

\begin{figure}
          \centering
           \input{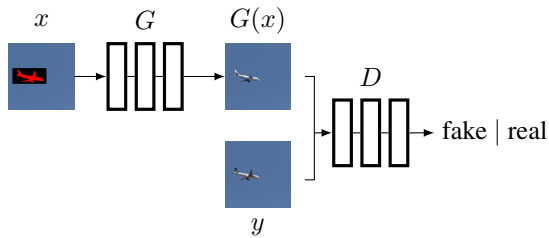}
           \caption{Proposed Pix2Pix Training. The generator and the discriminator are trained together. Thereby the generator learns to generate an image close to the ground truth $y$. Part of the generator input is replaced by a mask. The discriminator learns to decide whether an input is fake or real. \cite{pix2pix2017}}
           \label{fig:p2p-training}
        \end{figure}

    At training time the the \ac{cGAN} consists of two networks as shown in Figure \ref{fig:p2p-training}.
    The generator synthesizes images based on the provided inpainting image patch. The discriminator tries to differentiate between real and fake images.
    This classification is then used to train the generator.
    Given a dataset where the position and segmentation of objects in the images is known, we modify the input images by overlaying the segmentation mask which is marking object pixels. The direct surrounding neighborhood is colored uniformly black such that the whole inpainting mask has a rectangular shape. This image is used as input $x$ for the generator. Meanwhile, the original image from the dataset is used as ground truth $y$ for the discriminator. The generator now needs to create a new image with the desired object at the specified position, surrounded by a consistent background. Note that after the training, only the generator is used for data generation.
    
    % While the original GAN approach is capable of creating images, it has limited control what kind of image will be created. Later \acp{cGAN} added another input vector to the generator which allows to control the output image. This input vector can be an arbitrary number or input images. 
    % The discriminator needs to be adapted in such a way that it is capable of giving feedback based on the condition.
    % To achieve inpainting with GANs, one needs a conditional model that uses images instead of random and conditional vectors as input.
    % One well-known \ac{cGAN} is Pix2Pix, which can be used in a variety of image-to-image translations, such as from black-and-white to color images. Unlike the original GAN which only uses one set of training images to train the discriminator Pix2Pix needs two. An input image for the generator and a ground truth image for the discriminator for comparison \cite{cGAN}\cite{pix2pix2017}.
        
    %TODO: Add Losses?
        %\begin{itemize}
        %    \item GANs first introduced in 2014 (https://arxiv.org/abs/1406.2661)
        %    \item use two neural nets, Generator/ Discriminator
        %    \item later models with more functionality (conditional, Img2Img)
        %    \item Pix2Pix --> Img2Img GAN
        %    \item allows manipulation of images in flexible use cases (day2night, BW2Color, ...)
        %    \item Training Process different from original GAN (needs Input-Image and Ground Truth)
        %    \item Stefans pointcloud paper
        %\end{itemize}    

             %\begin{itemize}
             %    \item Pix2Pix without build-in Inpaint method
             %    \item idea: combine GT-Image with segmentation mask
             %    \item model learns to create objects based on masks
             %    \item similar to original Pix2Pix use cases, mayor difference the non masked part of the image needs to stay the same
             %\end{itemize}

    % Inpainting images with generative models allows masked image areas to be replaced with different content while keeping the rest unchanged. This can either be used for creating new content or removing undesired elements of the original image. We base our work up on the first principle which leverages the possibility to control the image synthesis process of a generative model. 
    % Given a four-cornered mask, a generative model is used to blend an object from a target domain class into a background image. The mask is later used to automatically derive ground truth labels.
    % The challenge is to create a natural-looking image so that the utility of the generated dataset is maximized.
    % The subsequent paragraphs describe the inpainting methods using Stable Diffusion and Pix2Pix.
    
    %\begin{itemize}
    %    \item generative models are capable of modeling conditional distributions (see https://arxiv.org/pdf/2112.10752.pdf)
    %    \item this allows to control the generation process (see https://arxiv.org/pdf/2112.10752.pdf)
    %\end{itemize}
\subsection{Object Synthesis using Stable Diffusion}\label{sec:synthesis-sd}
    In contrast to the Pix2Pix approach, we condition Stable Diffusion on the whole cropped inpainting patch. Additionally, a text prompt is provided that guides the type and appearance of the generated object. Unfortunately, with this approach, the exact size of the object cannot be controlled or is otherwise observable from the model output. The derived bounding box ground truth is set to the whole patch size which might be substantially larger than the actual object.

    The architecture of the \ac{LDM} consists of three main components.
    The first is an autoencoder that performs encoding and decoding from pixel space to a computationally efficient latent space. The second is a time-conditional UNet \cite{UNet} used as the backbone for the denoising process.
    The third and last is a domain-specific encoder that allows arbitrary conditioning of inputs, such as text prompts and images. The output embedding is then fed into the UNets via the cross-attention layer \cite{rombach2022high}.
    We build on the image-to-image functionality implemented in the Diffusers library\footnote{GitHub: https://github.com/huggingface/diffusers}. Because the existing pre-trained weights already generalize very well, we do not fine-tune Stable Diffusion on domain specific data in the context of this paper.
    % Our proposed inpainting method based on Stable Diffusion requires a four-cornered mask and a text prompt that guides what type of object gets synthesized. We build on the image-to-image functionality implemented in the Diffusers library\footnote{GitHub: https://github.com/huggingface/diffusers}. During inference, the masked area is cropped from the background image and resized to match the input size of the diffusion model. The diffusion process is then controlled by the given text prompt. The result is inserted back into the  background image. This method allows us to insert small objects into high-resolution images, which is especially important in the field of Airborne Object Detection, as mentioned earlier.

\section{Experiments}\label{sec:experiments}
In the following experiments, qualitative results for the inpainting process using Stable Diffusion and Pix2Pix are presented. Furthermore we show the dataset generation costs and investigate the usability of the generated dataset.

\subsection{Inpainting using Pix2Pix}
    For the experiments with Pix2Pix we use an open source PyTorch implementation\footnote{GitHub: https://github.com/junyanz/pytorch-CycleGAN-and-pix2pix} of the model.
    We train on a selection of about 1500 images from a proprietary dataset containing various airborne objects. Each frame is annotated with a bounding box, class label and a segmentation mask. No synthetic data of any kind is contained in the dataset.
    Table \ref{tab:p2p-hyperparams} contains the changed hyperparameters.
    We train for 2000 epochs on a resolution of 256x256 pixel. The Learning Rate Scheduling follows a cosine function starting at $2\mathrm{e}{-4}$ to $2\mathrm{e}{-6}$. As optimizer we use Adam, like the original implementation.

    \begin{table}
  \centering
  \begin{tabular}{@{}lc@{}}
    \toprule
    Hyperparameter & Value \\
    \midrule
    Resolution & 256x256\\
    Epochs & 2000\\
    Optimizer & Adam \\    
    Learning Rate & $2\mathrm{e}{-4}$ \\
    Learning Rate Scheduling & Cosine \\
    Final Learning Rate & $2\mathrm{e}{-6}$\\
    \bottomrule
  \end{tabular}
  \caption{Pix2Pix Hyperparameters. All hyperparameters that are not named are kept as in the original implementation.}
  \label{tab:p2p-hyperparams}
\end{table}
    
    The resulting images, shown in Figures \ref{fig:inpainting-b} and \ref{fig:inpainting-d}, prove that the proposed method works in general. Especially small objects are realistically looking. Problems occur if large segmentation masks are used. The coloring of the object itself is not realistic and around the object artefacts are visible, like shown in Figure \ref{fig:inpainting-b}. In some cases there is significant contrast between masked region and the rest of the patch. This is especially noticeable when the background image has an orange hue. This behaviour is probably due to training bias, while the used training data provides different backgrounds, sunsets are rather uncommon.
    When the inference result is merged with the large image, the cropped area is visible in some cases. Again this problem is particularly noticeable with orange backgrounds, see Figure \ref{fig:inpainting-b}.

    While the coloring of Pix2Pix still needs improvement, objects completely fill the given segmentation mask. This means that the position of the inpainted object is exactly at the specified location, which enables the creation of the required bounding boxes.
    
    %\begin{itemize}
    %    \item using Pix2Pix with UNet Generator, Resolution 256x256
    %    \item training on approximately 1500 Images, with one flying object each
    %    \item brief description of the dataset maybe an example image?
    %\end{itemize}

    %\begin{itemize}
    %    \item coloring not realistic
    %    \item shape is good 
    %    \item background recreation has some artefacts (especially for orange backgrounds)
    %    \item works also with more than one Object and segmentation masks not seen before
    %    \item best results with small objects and gray backgrounds (probably training bias)
    %\end{itemize}

%Not sure if we need this any more. Leave it as comment for now
%\subsection{Scaling Image Resolution}
%When using the discussed method to generate high-resolution images the problems from the inpainting persist. Notable is that only the segmentation mask of the cropped part is changed. Changes in the cropped part would be visible when looking at the whole image.
%    \begin{itemize}
%        \item Problems from the 256x256 images persist but not as notable
%        \item Objects need to be smaller than the crop
%    \end{itemize}

\subsection{Inpainting using Stable Diffusion}
Two high-resolution background images are used to evaluate the inpainting process with Stable Diffusion. We use one fixed inpainting mask per image for all experiments.
The prompts for guiding the diffusion process are kept simple and follow the scheme: "a photograph of [object], Nikon D850". The "[object]" part is replaced by the desired dataset class \eg "an airplane". The appendix "[...], Nikon D850" refers to a certain type of camera which results in a photo-realistic style of the synthesized image region. This particular model was chosen because it is a full-frame camera that is often used for landscape photography. 
The same set of hyperparameters presented in Table \ref{tab:sd-hyperparams} is used for all diffusion processes. They were selected by a grid search and a qualitative analysis of the results.

\begin{table}
  \centering
  \begin{tabular}{@{}lc@{}}
    \toprule
    Hyperparameter & Value \\
    \midrule
    Sampling Method & DDIM \\
    Sampling Steps & 50 \\
    Denoising Strength & 0.9\\
    CFG Scale & 7\\
    Resolution & 512x512\\
    \bottomrule
  \end{tabular}
  \caption{Stable Diffusion Hyperparameters. The same set of hyperparameters is used for all diffusion processes. All hyperparameters that are not named are kept as in the original implementation.}
  \label{tab:sd-hyperparams}
\end{table}

\begin{figure*}
  \centering
  \begin{subfigure}{.48\linewidth}
    \includegraphics[trim=1000px 1700px 1700px 0px,clip,width=\linewidth]{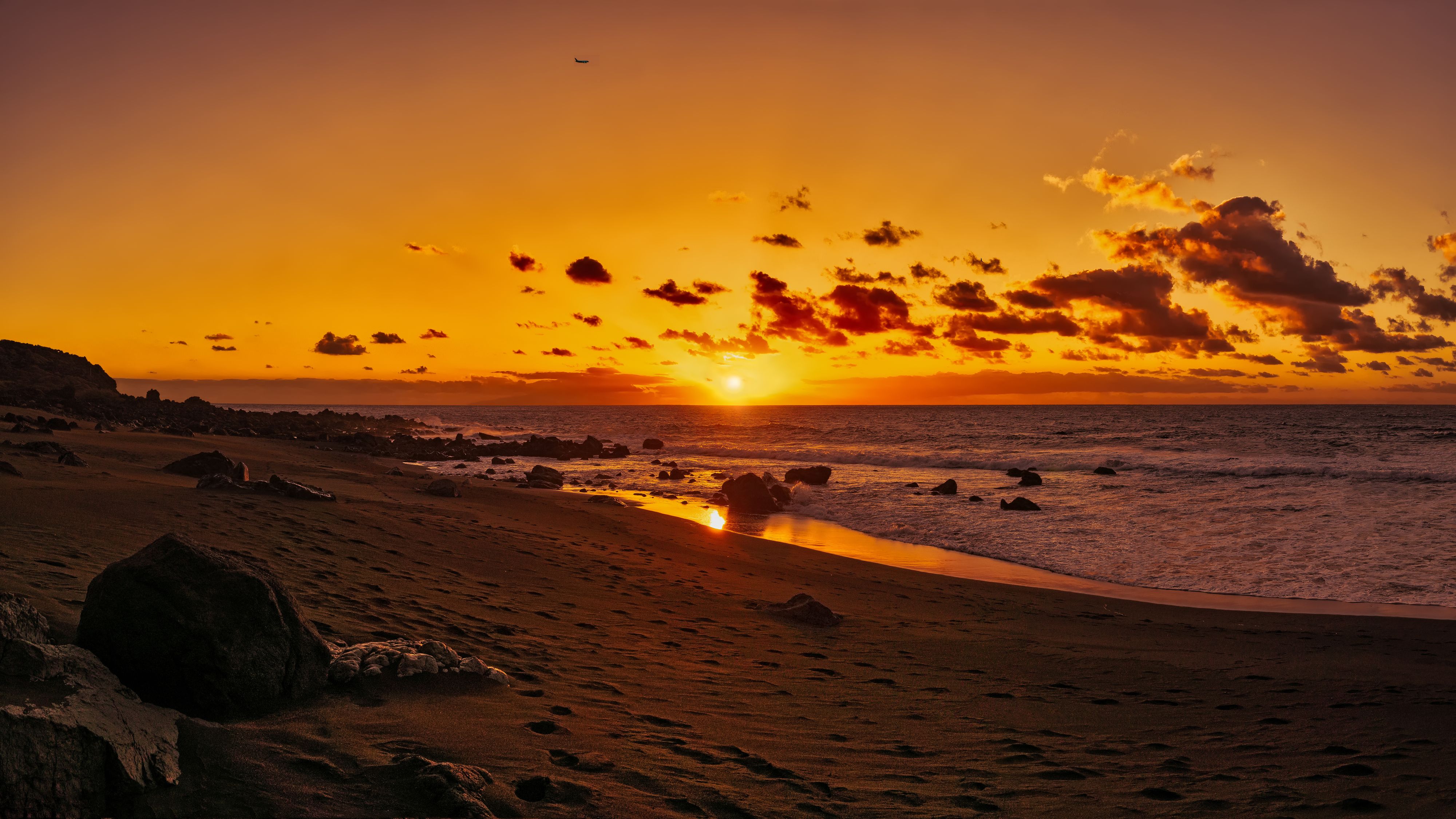}
    \caption{SD on Background 1\cite{background1}}
    \label{fig:inpainting-a}
  \end{subfigure}
  \hfill
  \begin{subfigure}{.48\linewidth}
    \includegraphics[trim=1000px 1700px 1700px 0px,clip,width=\linewidth]{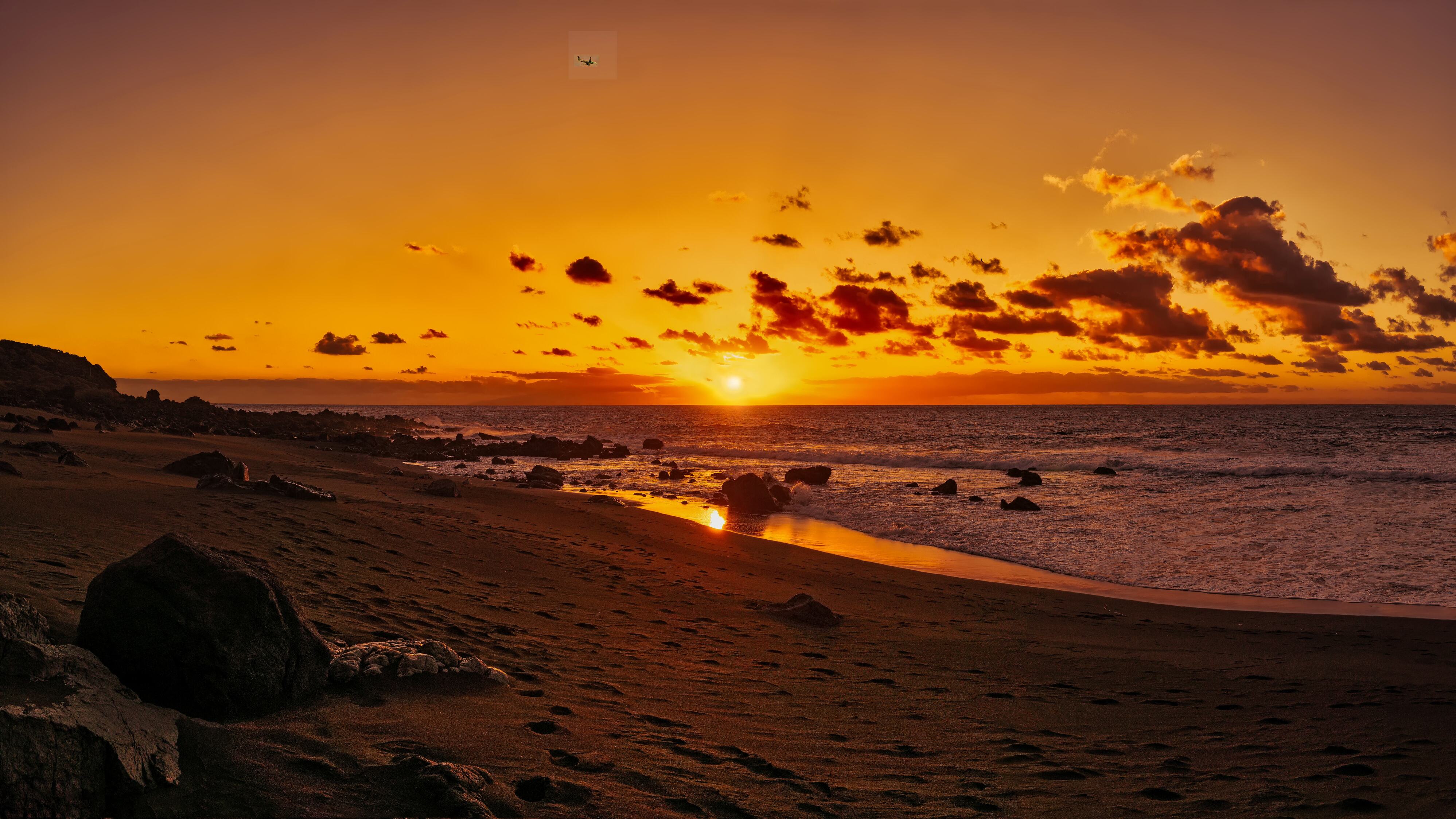}
    \caption{Pix2Pix on Background 1\cite{background1}}
    \label{fig:inpainting-b}
  \end{subfigure}
  \\
  \begin{subfigure}{.48\linewidth}
    \includegraphics[trim=2200px 1400px 500px 250px,clip,width=\linewidth]{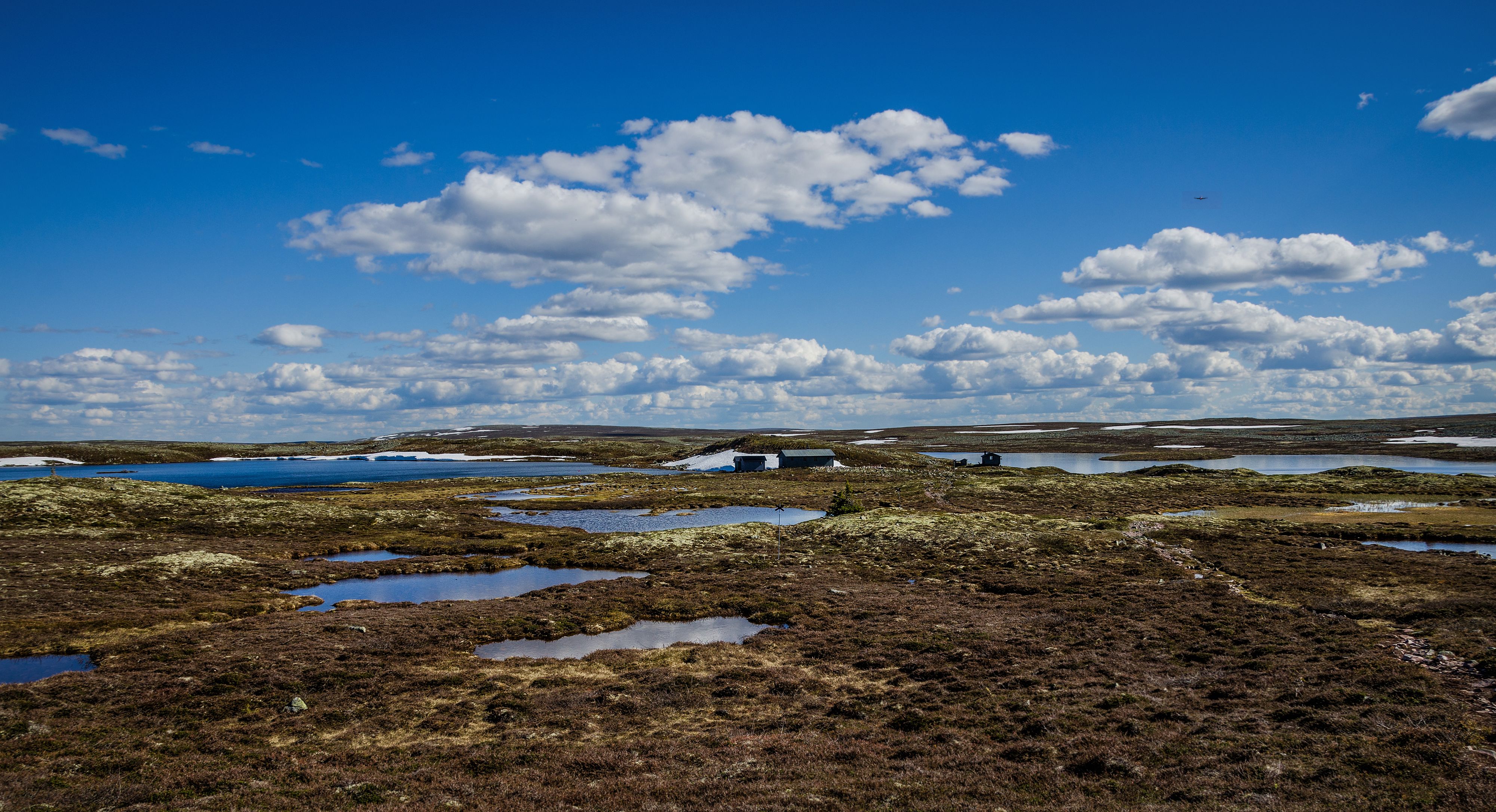}
    \caption{SD on Background 2\cite{background2}}
    \label{fig:inpainting-c}
  \end{subfigure}
  \hfill
  \begin{subfigure}{.48\linewidth}
    \includegraphics[trim=2200px 1400px 500px 250px,clip,width=\linewidth]{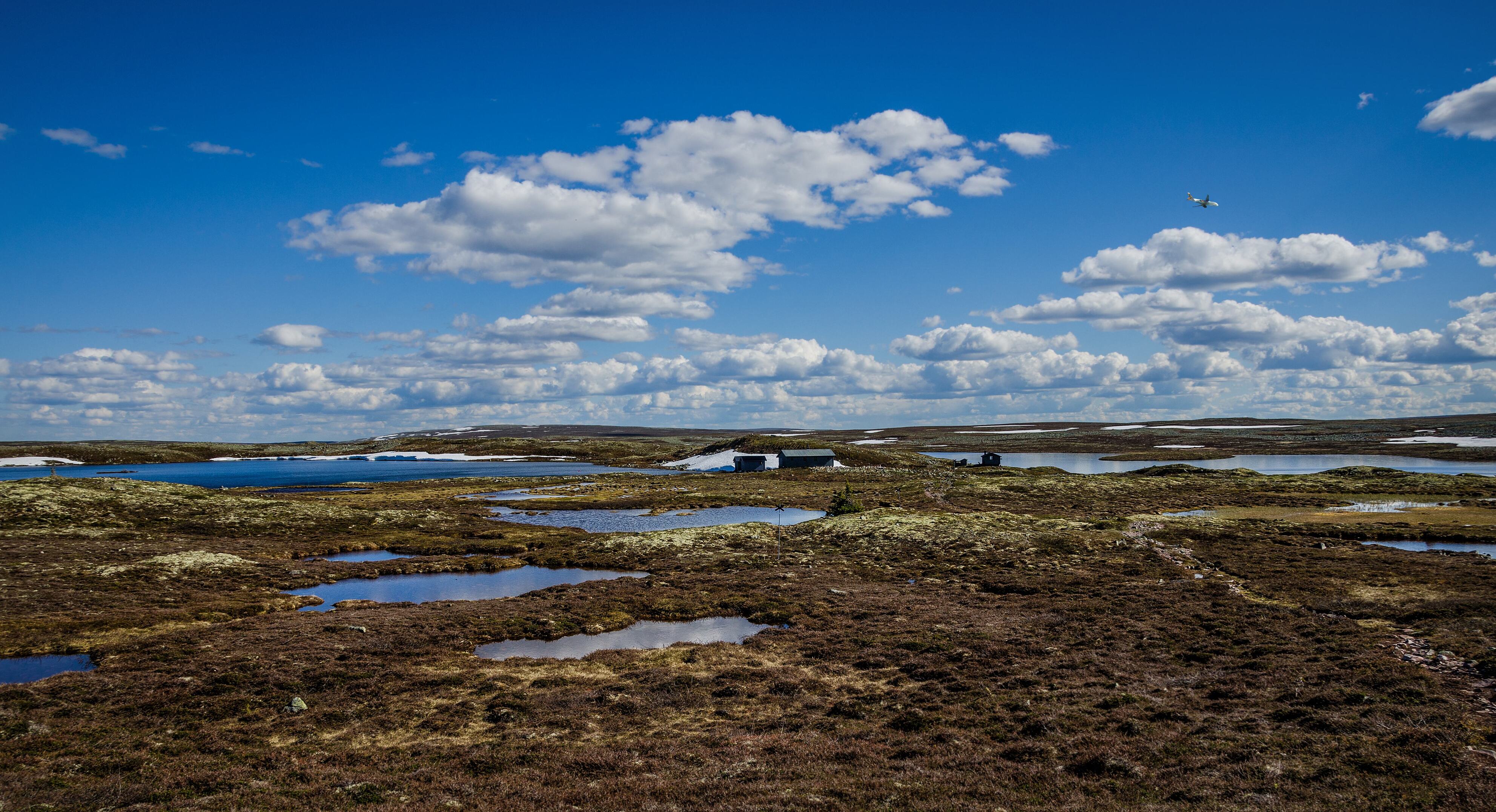}
    \caption{Pix2Pix on Background 2\cite{background2}}
    \label{fig:inpainting-d}
  \end{subfigure}
  \caption{Data Generation Results. The images show exemplary selected outputs of the proposed data generation pipeline. The left column is generated by using Stable Diffusion (SD) for inpainting. Pix2Pix was used for generating the right column.}
  \label{fig:inpainting}
\end{figure*}

Figures \ref{fig:inpainting-a} and \ref{fig:inpainting-c} show the inpainting results generated using the previously described method, prompt and hyperparameters. In general the generated objects look realistic and the synthetically generated region blends into the whole picture really well.
It can be clearly seen that the created objects do not completely fill the masked area. This fact makes it problematic to use the mask as a reference for the bounding box of the object. Apart from that, a slight color difference can be seen within the masked region. This could facilitate the detection of the pasted object.

\subsection{Dataset Generation}
%In order to indicate the usability of the generated data we synthesise a small dataset with 50 images. Figure \ref{fig:dataset} shows close-ups of some of the generated objects. Thereby the upper row shows qualitatively worse results compared to the lower row.
%As background we use 5 different images with resolution up to 7680x4320 pixels. Our trained Pix2Pix model is used as the generative model of the pipeline, because the extracted ground truth is more accurate than with Stable Diffusion. Furthermore the inference time is similar for both models, with the GAN approach requiring less VRAM during inference at about 1 GB. For comparison, Stable Diffusion requires about 8 GB.  
%Using Pix2Pix with an input size of 256x256, the time to create a single image with the proposed pipeline is about 0.5 seconds. The measurement was performed with a NVIDIA RTX 2060 with 6 GB video memory and an Intel i5-9600K@3.7GHz. A large scale dataset with \eg $10^5$ frames could be generated within 14 hours if executed single threaded. Note that the implementation hasn't been optimized for runtime performance and the task is highly parallelizable.
%In the scope of this paper we limit the number of inpainted objects per image to $1$ because for the \ac{DAA} use case multiple encounters at once are quite unlikely in most airspaces. Theoretically the proposed method could also be adapted to allow multiple objects.
In order to study the usability of the generated data, we synthesise a dataset containing about 7000 images using different backgrounds. The background images cover different weather, cloud and lighting, an example of this can be seen in Figure \ref{fig:inpainting}. Further no preexisting objects should appear in the image and at least the top half of the image has to show sky. The size and aspect ratio of the images can vary but should at least be bigger than the crop used for inpainting.
The airborne objects are sampled random from the dataset used to train Pix2Pix. This results in a matching class distribution of the generated dataset and the training dataset. The number of instances per class can be seen in Table \ref{tab:synthetic_dataset_classes}.
Our trained Pix2Pix model is used as the generative model of the pipeline, because the extracted ground truth is more accurate than with Stable Diffusion. Furthermore the inference time is similar for both models, with the GAN approach requiring less VRAM during inference at about 1 GB. For comparison, Stable Diffusion requires about 8 GB. This low video memory requirement enables our proposed data generation pipeline to run on consumer hardware.

\begin{figure*}
  \centering
  \begin{subfigure}{.20\linewidth}
    \includegraphics[height=0.7\linewidth, width=\linewidth]{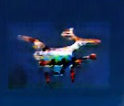}
    \caption{Drone}
    \label{fig:dataset-a}
  \end{subfigure}
  \hfill
  \begin{subfigure}{.20\linewidth}
    \includegraphics[height=0.7\linewidth, width=\linewidth]{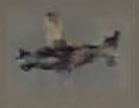}
    \caption{Small Airplane}
    \label{fig:dataset-b}
  \end{subfigure}
  \hfill
  \begin{subfigure}{.20\linewidth}
    \includegraphics[height=0.7\linewidth, width=\linewidth]{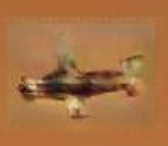}
    \caption{Small Airplane}
    \label{fig:dataset-c}
  \end{subfigure}
  \hfill
  \begin{subfigure}{.20\linewidth}
    \includegraphics[height=0.7\linewidth, width=\linewidth]{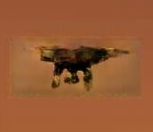}
    \caption{Drone}
    \label{fig:dataset-d}
  \end{subfigure}
  \\
  \begin{subfigure}{.20\linewidth}
    \includegraphics[height=0.7\linewidth, width=\linewidth]{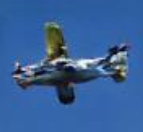}
    \caption{Small Airplane}
    \label{fig:dataset-e}
  \end{subfigure}
  \hfill
  \begin{subfigure}{.20\linewidth}
    \includegraphics[height=0.7\linewidth, width=\linewidth]{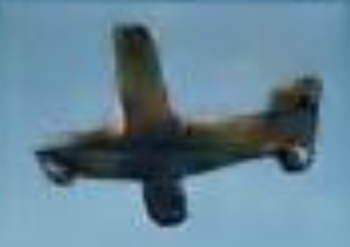}
    \caption{Small Airplane}
    \label{fig:dataset-f}
  \end{subfigure}
  \hfill
  \begin{subfigure}{.20\linewidth}
    \includegraphics[height=0.7\linewidth, width=\linewidth]{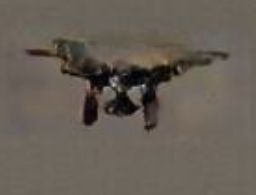}
    \caption{Drone}
    \label{fig:dataset-g}
  \end{subfigure}
  \hfill
  \begin{subfigure}{.20\linewidth}
    \includegraphics[height=0.7\linewidth, width=\linewidth]{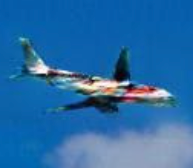}
    \caption{Airplane}
    \label{fig:dataset-h}
  \end{subfigure}
  \caption{Data Generation Results. The images show close-ups of selected samples of the generated data set. The classes of the objects are mentioned in the captions. The upper row shows qualitatively worse results compared to the lower row.}
  \label{fig:dataset}
\end{figure*}

Using Pix2Pix with an input size of 256x256, the time to create a single image with the proposed pipeline is about 0.5 seconds. This measurement was performed on a NVIDIA RTX 2060 and an Intel i5-9600K@3.7GHz.
Some examples of the generated images can be seen in Figure \ref{fig:dataset}.
For our small scale test we generated 5900 training images as well as 590 test and validation images. Each split uses different background images. In order to achieve a high variety of objects the training split contains the 4300 objects used in training Pix2Pix while the other two splits share 500 objects. However those objects where never seen by Pix2Pix.
The objects are placed in the upper half of the image, to make sure they appear in the sky and not on the ground. 
Only one object is inpainted per image in the limited scope of this paper. The proposed method is easily extendable to multiple objects per image.
%Doesen't fit the page
%\begin{table}[]
%    \centering
%    \begin{tabular}{c|c|c|c|c}
%         Class & Number of Instances over all Splits & train split & validation split & test split \\
%         large airplane & 1695 & 1416 & 142 & 137\\
%         small airplane & 1255 & 1046 & 96 & 113\\
%         very small airplane & 46 & 38 & 2 & 6\\
%         helicopter & 2201 & 1812 & 206 & 183\\
%         drone & 961 & 800 & 86 & 75\\
%         hot air balloon & 315 & 268 & 21 & 26\\
%         paraglider & 565 & 492 & 32 & 41\\
%         airship & 42 & 28 & 5 & 9\\
%         UFO & 0 & 0 & 0 & 0
%    \end{tabular}
%    \caption{Synthetic Dataset Overview}
%    \label{tab:synthetic_dataset_classes}
%\end{table}

\begin{table}[]
    \centering
    \begin{tabular}{@{}lc@{}}
        \toprule
         Class & Number of Instances\\
         \midrule       
         Large Airplane & 1695 \\
         Small Airplane & 1255 \\
         Very Small Airplane & 46 \\
         Helicopter & 2201 \\
         Drone & 961 \\
         Hot Air Balloon & 315 \\
         Paraglider & 565 \\
         Airship & 42 \\
         \bottomrule
    \end{tabular}
    \caption{Synthetic Dataset Overview over all splits. Here one image contains one instance.}
    \label{tab:synthetic_dataset_classes}
\end{table}

\begin{table}
  \centering
  \begin{tabular}{@{}lc@{}}
    \toprule
    Hyperparameter & Value \\
    \midrule
    Resolution & 256x256\\
    Epochs & 300\\
    Optimizer & SGD \\    
    Batch Size & 64 \\ 
    Learning Rate & $1\mathrm{e}{-2}$ \\
    Learning Rate Scheduling & Cosine \\
    Final Learning Rate & $1\mathrm{e}{-4}$\\
    \bottomrule
  \end{tabular}
  \caption{YOLOv8s Hyperparameters. All hyperparameters that are not named are kept as in the original implementation.}
  \label{tab:yolov8s-hyperparams}
\end{table}

\textbf{Domain Gap Investigation}
    To investigate the domain gap, we evaluate the object detection performance on the synthetically generated data and the dataset used for training the Pix2Pix model. We train an open source adaption\footnote{GitHub: https://github.com/ultralytics/ultralytics} of the YOLO \cite{redmon2018yolov3} object detector only on the real world dataset.
    Using $80\%$ of the dataset for training and 20\% for validation, a validation mAP of $70.1\%$ could be achieved. We then evaluated the model performance on the synthetic dataset. Thereby a mAP of $32.9\%$ was reached. All tracked validation metrics are listed in Table \ref{tab:od-results}. Note that the backgrounds used for generating images are unseen for both the generator and the object detector. 
    
    \begin{table}
      \centering
      \begin{tabular}{@{}l c c c c@{}}
        \toprule
        Dataset & mAP & mAP@50 & Precision & Recall \\
        \midrule
        Inhouse & 0.701 & 0.805 & 0.866 & 0.654\\
        Generated & 0.329 & 0.600 & 0.542 & 0.713\\
        \bottomrule
      \end{tabular}
      \caption{Object Detection Validation Results. The table compares common object detection metrics for the inhouse dataset and the generated one. The recall is on par for both but the classification performance is decreased on the generated dataset.}
      \label{tab:od-results}
    \end{table}
    
    The decrease in detection performance can be explained if precision and recall are considered. The recall is on a similar level for both datasets but the precision is significantly decreased on the generated dataset compared to the real validation set. This leads to the conclusion that the objects are detected successfully but are misclassified more frequently on synthetic data. A likely explanation for that are the abstract textures that the generated objects tend to have. 

\begin{figure}
      \centering
      \begin{subfigure}{.48\linewidth}
        \includegraphics[trim=50px 720px 1400px 0px,clip,width=\linewidth]{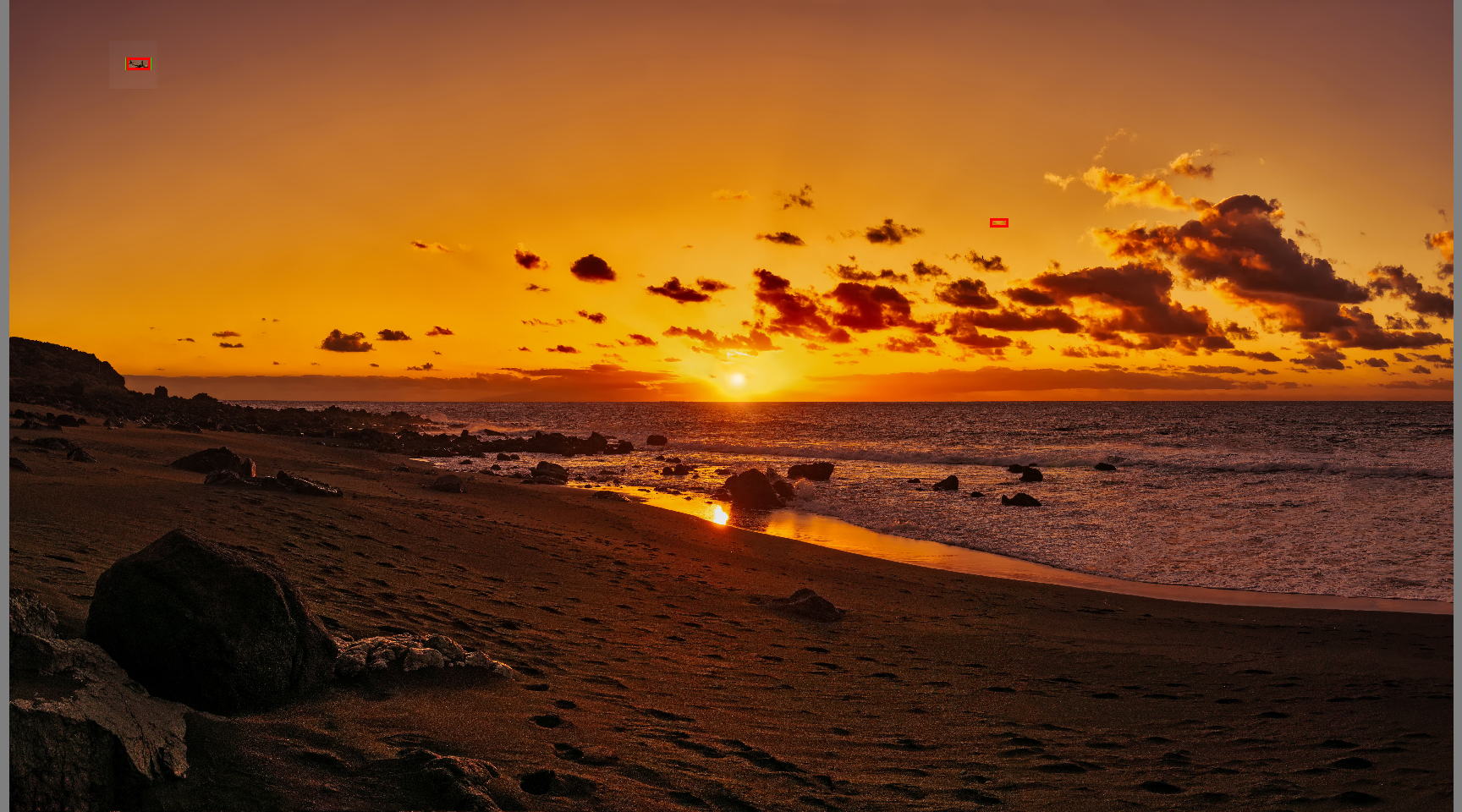}
        \caption{OD, Background 1 \cite{background1}}
        \label{fig:od-result-a}
      \end{subfigure}
      \hfill
      \begin{subfigure}{.48\linewidth}
        \includegraphics[trim=50px 500px 1200px 50px,clip,width=\linewidth]{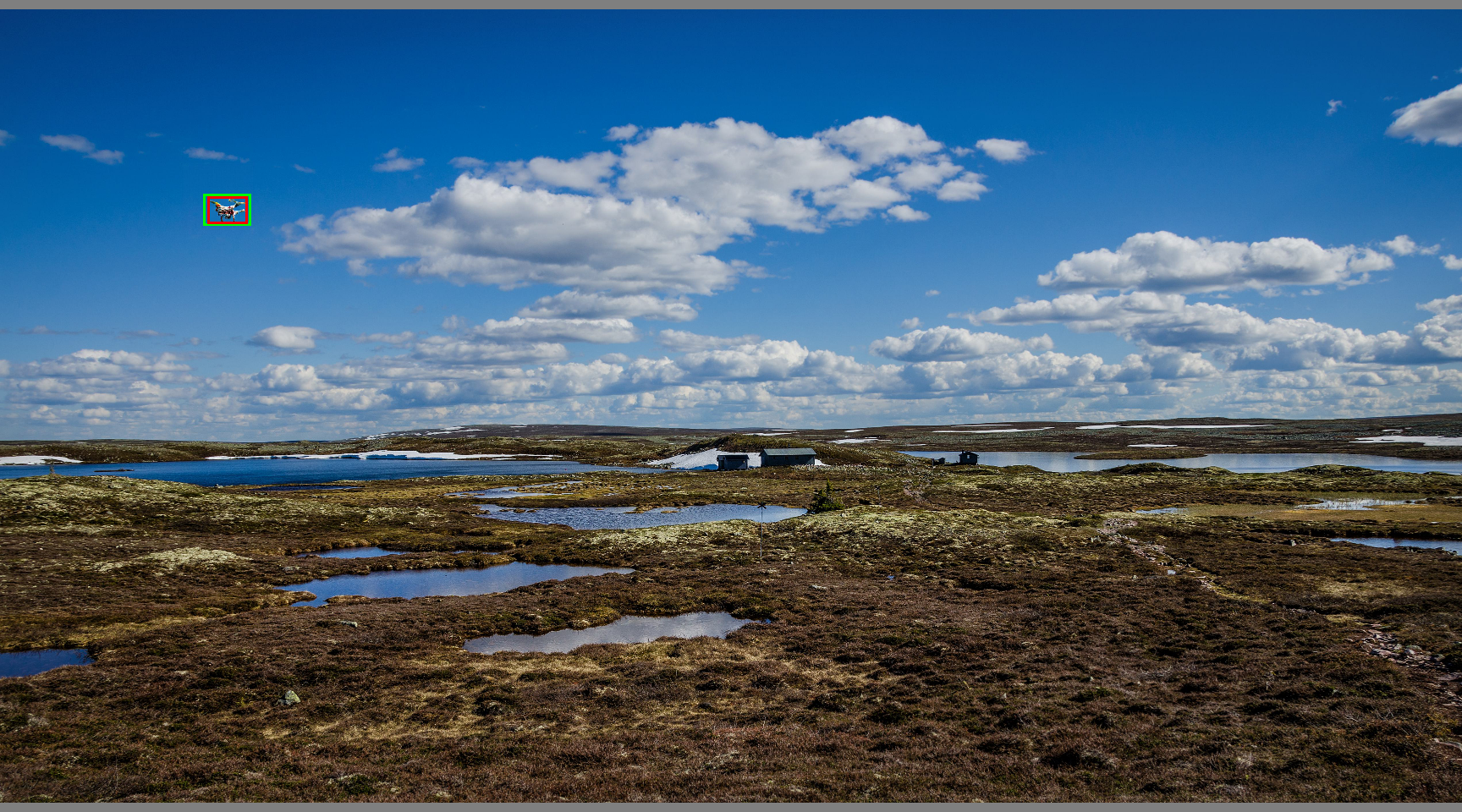}
        \caption{OD, Background 2 \cite{background2}}
        \label{fig:od-result-b}
      \end{subfigure}
      \caption{Object Detection Results. Both images show a successful object detection on generated images. The predicted bounding box nearly matches the ground truth.}
      \label{fig:od-result}
    \end{figure}

    The above statement can be confirmed by visualizing the output of the object detection model, as shown in Figure \ref{fig:od-result}. The ground truth box is drawn in green. The red box is the predicted object position. It is noticeable that the color difference within the inpainted area does not seem to negatively affect the prediction. 
    In general, the observations show that there is room for improvement in closing the domain gap, although the object recognition model may be successfully deceived in some images. This shows that the proposed data generation pipeline is suitable for generating large training datasets. However, the quality of the generated objects needs further improvement.

\section{Conclusion}
   %\begin{itemize}
   %    \item Diffusion works best for generic image tasks - without fine-tuning
   %    \item problems occur when the task gets more specific -> fine-tuning 
   %    \item Recent work on LDMs (examples) could solve these problems
   %    \item Layout-to-image synthesis shown in LDM paper not implemented yet, perfect approach for object detection
   % \end{itemize}
   % \begin{itemize}Das quadrat ist hier ein unwichtiges 
   %     \item Pix2Pix can learn to differentiate between background and segmentation mask
   %     \item can learn to fill in segmentation and recreate the background around the object
   %     \item Training needs fine-tuning (especially coloring and orange backgrounds)
   %     \item inpainting works well, masks get filled in while the background stays the same
   %     \item creating large images is fast enough
   % \end{itemize}

    % draft:
    We tested two different generative models for inpainting airborne objects into high-resolution images, with the goal of creating datasets for training and validation of object detection models. While diffusion models show good qualitative results out of the box, the extracted ground truth bounding boxes have limited accuracy. However, for the training and validation of object detection models, high quality ground truth is essential. To overcome this issue, significant adaptions to the \ac{LDM} and a fine-tuning might be necessary.
    
    The experiments with the trained Pix2Pix model show its capabilities of inpainting objects into existing images.
    While the coloring of the synthesized image region is a problem, the position of the objects is exactly at the given location and the mask is completely filled out. That allows for accurate extraction of ground truth bounding boxes.
    
    We have shown that our proposed data generation pipeline allows to create \ac{DAA} datasets of high-resolution images in feasible time. It is possible to swap out the generative model as needed. Thus, the method can be easily applied to future state-of-the-art generative models. In future work it needs to be demonstrated whether training an object detector on the generated data has an actual benefit on the resulting performance, robustness and generalizability. Possible approaches would be pre-training or enrichment of the main dataset. Extension to simultaneous generation of multiple objects or other tasks such as semantic segmentation are other interesting areas of research.

{\small
\bibliographystyle{ieee_fullname}
\bibliography{egbib}
}

\end{document}